\documentclass[lettersize,journal]{IEEEtran}
\usepackage{amsmath,amsfonts}
\usepackage{algorithmic}
\usepackage{algorithm}
\usepackage{array}
\usepackage[caption=false,font=normalsize,labelfont=sf,textfont=sf]{subfig}
\usepackage{textcomp}
\usepackage{stfloats}
\usepackage{url}
\usepackage{verbatim}
\usepackage{graphicx}
\usepackage{cite}
\usepackage{bm}
\usepackage{amssymb}
\usepackage{xcolor}
\usepackage{subcaption}
\usepackage{tikz}
\usetikzlibrary{shapes.geometric, arrows.meta, positioning, calc}
\usepackage{hyperref}
\usepackage{soul}
\usepackage{orcidlink}
\hypersetup{colorlinks=true, citecolor=blue, linkcolor=red, urlcolor=blue}

\begin{document}

\title{Strain-Parameterized Coupled Dynamics and Dual-Camera Visual Servoing for Aerial Continuum Manipulators}

\author{Niloufar Amiri and Farrokh Janabi-Sharifi
\thanks{This work was supported through Natural Sciences and Engineering Research Council of Canada under grant number 2023-05542, and National Research Council Canada under grant number AI4L-128-1.}%
\thanks{The authors are with the Department of Mechanical, Industrial, and Mechatronics Engineering,
Toronto Metropolitan University, ON, Canada.
E-mail: niloufar.amiri@torontomu.ca (corresponding author); 
fsharifi@torontomu.ca.}%

}



\maketitle

\begin{abstract}
Tendon-driven aerial continuum manipulators (TD-ACMs) combine the maneuverability of uncrewed aerial vehicles (UAVs) with the compliance of lightweight continuum robots (CRs). Existing coupled dynamic modeling approaches for TD-ACMs incur high computational costs and do not explicitly account for aerial platform underactuation. To address these limitations, this paper presents a generalized dynamic formulation of a coupled TD-ACM with an underactuated base. The proposed approach integrates a strain-parameterized Cosserat rod model with a rigid-body model of the UAV into a unified Lagrangian ordinary differential equation (ODE) framework on $\mathrm{SE}(3)$, thereby eliminating computationally intensive symbolic derivations. Building upon the developed model, a robust dual-camera image-based visual servoing (IBVS) scheme is introduced. The proposed controller mitigates the field-of-view (FoV) limitations of conventional IBVS, compensates for attitude-induced image motion caused by UAV lateral dynamics, and incorporates a low-level adaptive controller to address modeling uncertainties with formal stability guarantees. Extensive simulations and experimental validation on a compact custom-built prototype demonstrate the effectiveness and robustness of the proposed framework in real-world scenarios.
\end{abstract}


\section{Introduction}
Uncrewed aerial manipulators (UAMs) extend the capabilities of conventional uncrewed aerial vehicles (UAVs) by enabling physical interaction with the environment. Early UAM platforms predominantly employed rigid-link aerial manipulators (RA-AMs), which benefit from well-established rigid-body modeling and control methodologies \cite{ollero2021past}. However, their limited compliance restricts safe and robust operation in uncertain or contact-rich environments.

To overcome these limitations, aerial continuum manipulators (ACMs) have emerged as lightweight and compliant alternatives \cite{samadikhoshkho2020modeling, jalali2022aerial}. In tendon-driven ACMs (TD-ACMs), the manipulator exhibits distributed deformation, configuration-dependent stiffness, and strong bidirectional coupling between strain and applied forces. Although compliance enhances safety and adaptability, it significantly complicates the modeling, state estimation, and control of TD-ACMs, leaving several fundamental challenges unresolved.

\subsection{Modeling}

Mechanical models are essential for model-based control design and simulation-based analysis. While kinematic and quasi-static formulations \cite{amiri2025high, peng2023aecom} provide useful approximations, they are unable to capture the highly nonlinear behavior of TD-ACMs and the dynamic coupling between UAV motion and continuum robot (CR) deformation. The primary challenge lies in incorporating the distributed dynamics of the CR and its strong bidirectional coupling with the underactuated UAV platform \cite{tummers2023cosserat}.

Unlike RA-AMs, where link inertias and joint coordinates are explicitly defined, CRs exhibit distributed curvature and strain, resulting in effectively infinite degrees of freedom (DoFs), configuration-dependent compliance, and distributed internal forces. These characteristics render simplified geometric models, such as those based on constant-curvature (CC) approximations \cite{della2018dynamic} or classical beam theories \cite{liu2021effect}, insufficient accurately predicting CR dynamics under large deflections, external loading, and dynamic flight conditions. More accurate formulations, such as lumped-mass and Cosserat rod models \cite{tummers2023cosserat, renda2020geometric, boyer2020dynamics}, often incur prohibitive computational costs. 

Existing approaches typically adopt decoupled modeling strategies, treating the UAV–CR interaction as a disturbance \cite{samadikhoshkho2020modeling, chien2023design, hashemi2023robust}. This assumption breaks down when the manipulator inertia is non-negligible, during load transportation tasks, or under aggressive maneuvers. Fully coupled models have been proposed using finite-dimensional Lagrangian formulations \cite{samadikhoshkho2022vision, samadikhoshkho2021coupled} or augmented rigid-body representations \cite{szasz2022modeling}; however, these approaches either rely on the CC assumption or suffer from computational inefficiency.

To the best of the authors’ knowledge, a fully coupled, geometrically accurate, and computationally efficient dynamic model for TD-ACMs has not yet been reported in the literature.

\subsection{Low-Level Model-Based Control}

A control framework capable of handling uncertainties is essential for reliable operations of TD-ACMs. They are subject to substantial uncertainties arising from unmodeled dynamics, backbone material variability, tendon friction and hysteresis, and aerodynamic disturbances. Furthermore, the underactuation of commonly employed UAVs introduces additional coupling between translational and rotational dynamics. These factors, combined with strong UAV–CR coupling, render kinematic \cite{chien2021kinematic, peng2023aecom} and decoupled dynamic control strategies \cite{samadikhoshkho2020modeling, hashemi2023robust} inadequate for achieving robust performance.

Existing coupled approaches often rely on simplified CC-based models \cite{samadikhoshkho2022vision, samadikhoshkho2021coupled, szasz2022modeling}, which limit their accuracy under large deflections, distributed external loading, and high-speed maneuvers. Furthermore, fully coupled control methods remain largely confined to simulation-based validation due to the absence of computationally tractable dynamic models.

In summary, TD-ACMs require coupled, model-based dynamic control frameworks that balance accuracy and computational efficiency while ensuring robustness to system and environmental uncertainties. Such unified solutions, however, remain an open challenge.

\subsection{High-level Visual Servo Control}

While low-level dynamic control can mitigate TD-ACM parametric uncertainties, successful aerial manipulation ultimately depends on reliable perception and task-level regulation, for instance, to track a target with uncertain or time-varying position. For localization, conventional UAM systems often rely on motion-capture systems, the global positioning system (GPS), or simultaneous localization and mapping (SLAM), whose performance may degrade in GPS-denied, cluttered, or feature-poor environments. In the absence of joint encoders and explicit forward kinematic formulations, these challenges are further amplified in TD-ACMs due to indirect and uncertain state estimation of the CR.

Visual servoing (VS) offers a compelling alternative by reducing reliance on global localization systems, enabling direct regulation of the end-effector pose (e.g., relative to the target), and improving robustness to calibration errors, particularly in image-based visual servoing (IBVS) \cite{chaumette2006visual, janabi2010comparison}. For TD-ACMs, VS is especially attractive because it circumvents the need for precise tip-state reconstruction from uncertain strain measurements. However, VS design for TD-ACMs differs fundamentally from that for RA-AMs. In eye-in-hand (EIH) configurations, for instance, the camera pose depends on the deformable continuum shape, resulting in a more complex and nonlinear image feature mapping \cite{santamaria2019visual}. Moreover, CR flexibility may induce image distortion during motion \cite{nazari2022visual}, while frequent occlusions and field-of-view (FoV) loss further complicate control. Eye-to-hand (ETH) configurations improve global visibility but typically reduce spatial resolution and maneuverability \cite{ubellacker2024high, sato2025development, zhong2019practical}.

Dual-camera architectures, combining EIH and ETH configurations, leverage complementary advantages by enabling supervisory FoV management and high-resolution feedback for accurate local tracking. While dual-camera VS has been applied to rigid manipulators \cite{cuevas2018hybrid} and fixed-base CRs \cite{albeladi2022hybrid, zhang2021image}, it has not yet been developed for TD-ACMs.

\subsection{Approach and Contributions}

This work addresses the aforementioned challenges through an integrated vision-based control framework for TD-ACMs. The main contributions are summarized as follows:

\begin{itemize}
    \item A unified, multi-level dual-camera vision-based control framework for robust TD-ACM operation, integrating a high-level IBVS scheme with a low-level adaptive coupled controller.

    \item A dual-camera VS architecture that leverages the robustness of IBVS to modeling errors, mitigates FoV loss, and compensates for UAV underactuation directly in the image space.

    \item An adaptive, model-based coupled controller with formal stability guarantees, enabling accurate low-level regulation under CR parametric uncertainties and strong UAV-CR coupling.

    \item A geometrically exact, strain-parameterized coupled dynamic model on $\text{SE}(3)$ that, unlike existing Lagrangian formulations \cite{samadikhoshkho2021coupled}, preserves CR stiffness and damping, accounts for distributed body wrenches, and enables efficient computation without requiring computationally intensive symbolic derivations.

    \item Extensive simulation and experimental validation on a custom TD-ACM platform, demonstrating robustness to parametric uncertainties and environmental disturbances, such as mechanical vibrations.
\end{itemize}

\section{Strain-Parameterized Coupled Modeling}

\begin{figure}[t!]
    \centering
    \includegraphics[width=1\linewidth]{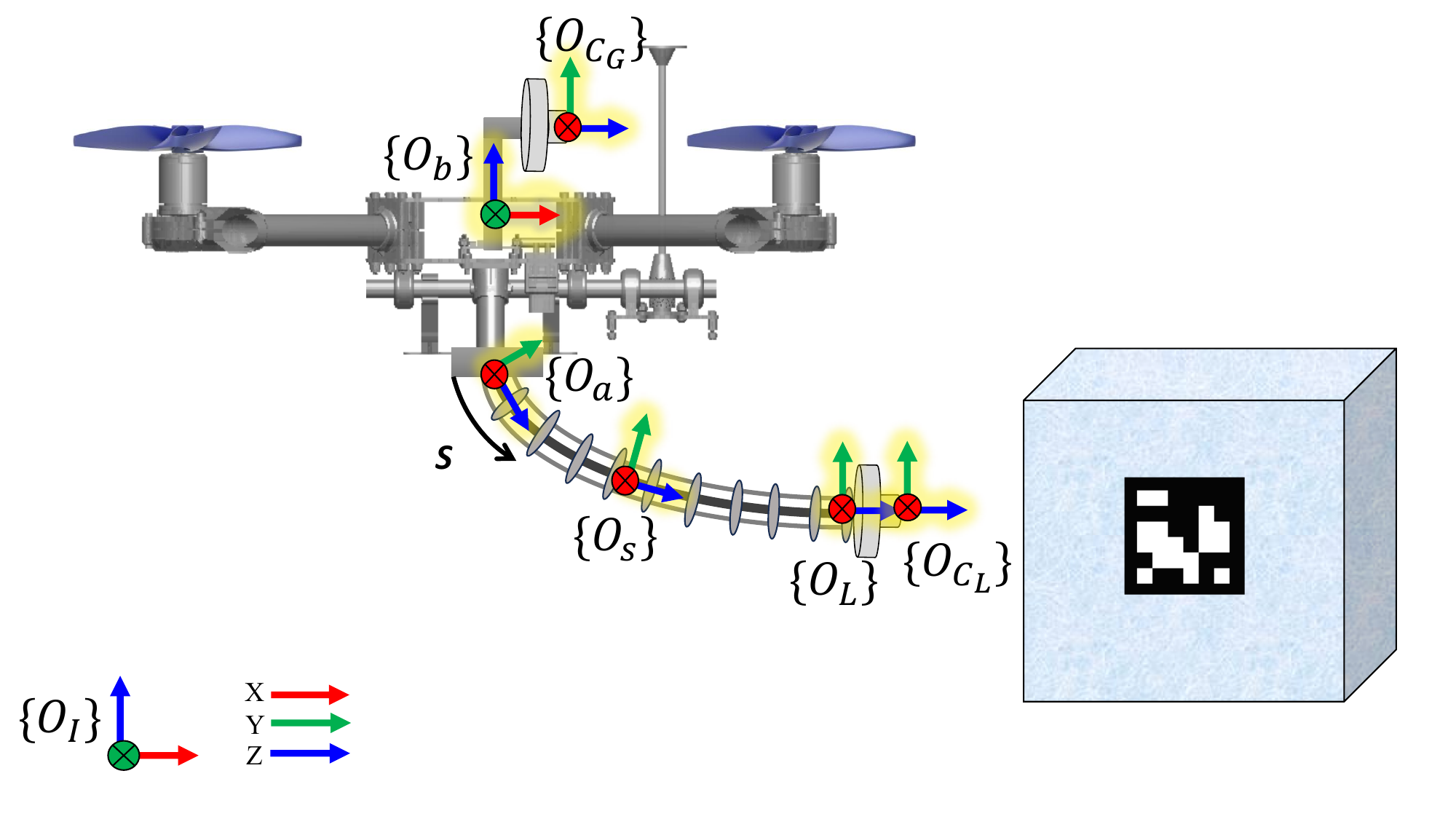}
    \caption{Schematic of the TD-ACM and key coordinate frames.}
    \label{fig:coordinates}
\end{figure}

In the inertial frame with origin $O_I$ depicted in Fig.~\ref{fig:coordinates}, consider a UAV with position vector $\bm{r}_b = [x, y, z]^\top \in \mathbb{R}^3$ and Euler angles $\bm{\Phi}_b = [\phi, \theta, \psi]^\top \in \mathbb{R}^3$. The homogeneous configuration of the UAV on the special Euclidean group $\mathrm{SE}(3)$ is
\begin{equation}
\bm{g}_b =
\begin{bmatrix}
\bm{R}_b & \bm{r}_b \\
\bm{0}_{1 \times 3} & 1
\end{bmatrix}
\in \mathrm{SE}(3),
\label{equation:0}
\end{equation}
where $\bm{R}_b \in \mathrm{SO}(3)$ is the rotation matrix of the UAV frame $O_b$, defined using the ZYX Euler-angle convention as $\bm{R}_b = \bm{R}_z(\psi)\bm{R}_y(\theta)\bm{R}_x(\phi)$. Throughout this paper, omitting the superscript in $A_a^{I}$ indicates that the quantity is expressed in the inertial frame $O_I$.

The CR mounted on the platform is characterized by the strain-based configuration variables
$\bm{q}_s = [q_{s_1}, \dots, q_{s_{n_s}}]^\top \in \mathbb{R}^{n_s}$,
where $n_s$ denotes the number of CR DoFs. The generalized coordinate vector of the whole-body TD-ACM is defined as
\begin{equation}
\bm{q} \triangleq [\phi, \theta, \psi, x, y, z, q_{s_1}, \dots, q_{s_{n_s}}]^\top \in \mathbb{R}^n,
\end{equation}
where $n = 6 + n_s$. The Jacobian of the UAV expressed in $O_b$ is
\begin{equation}
\bm{J}_{b}^b(\bm{q}) =
\begin{bmatrix}
\bm{T} & \bm{0}_{3 \times 3} & \bm{0}_{3 \times n_s} \\
\bm{0}_{3 \times 3} & \bm{R}_b^\top & \bm{0}_{3 \times n_s}
\end{bmatrix}
\in \mathbb{R}^{6\times n},
\end{equation}
where $\bm{T}$ maps the Euler-angle rates to the UAV angular velocity expressed in $O_b$ under the ZYX convention:
\begin{equation}
\bm{T} =
\begin{bmatrix}
1 & 0 & -\sin\theta \\
0 & \cos\phi & \sin\phi \cos\theta \\
0 & -\sin\phi & \cos\phi \cos\theta
\end{bmatrix}
\in \mathbb{R}^{3\times3}.
\label{equation:11}
\end{equation}
Accordingly, the UAV twist expressed in $O_b$ is $\bm{v}_b^{b} = \bm{J}_b^b(\bm{q}) \dot{\bm{q}}$.

The homogeneous transformation of a CR cross-section, expressed in the inertial frame and parameterized by the arc-length coordinate $s \in [0,L]$ measured from the attachment frame $O_a$, is defined as
\begin{equation}
\bm{g}_s = \bm{g}_b \bm{g}_{a}^{b} \bm{g}_{s}^{a} =
\begin{bmatrix}
\bm{R}_s & \bm{r}_s \\
\bm{0}_{1 \times 3} & 1
\end{bmatrix}
\in \mathrm{SE}(3),
\label{equation:6}
\end{equation}
where $\bm{g}_{a}^{b}$ is the constant transformation from the attachment frame $O_a$ to the UAV frame $O_b$, $\bm{g}_{s}^{a}$ is the configuration of the CR cross-section relative to $O_a$, and $L$ is the total length of the CR. Here, $\bm{R}_s$ and $\bm{r}_s$ denote the rotation matrix and position vector at $s$, respectively.

The derivative of the relative configuration is given by
\begin{equation}
\bm{g}^{a \prime}_s = \bm{g}^a_s \widehat{\bm{\xi}},
\label{equation:12}
\end{equation}
where $(\cdot)^\prime$ denotes differentiation with respect to $s$, and $\widehat{\bm{\xi}} \in \mathfrak{se}(3)$ is the strain twist:
\begin{equation}
\widehat{\bm{\xi}} =
\begin{bmatrix}
\operatorname{S}(\bm{k}) & \bm{u} \\
\bm{0}_{1 \times 3} & 0
\end{bmatrix} \in \mathfrak{se}(3),
\label{equation:13}
\end{equation}
where $\bm{k} = [k_1, k_2, k_3]^\top \in \mathbb{R}^3$ and $\bm{u} = [u_1, u_2, u_3]^\top \in \mathbb{R}^3$ are the rotational and linear components of the strain twist, respectively. The exponential solution of \eqref{equation:12} is obtained by integrating along $s$:
\begin{equation}
\bm{g}^a_s = \exp\!\left( \widehat{\bm{\Omega}}(s) \right),
\label{equation:18}
\end{equation}
where $\widehat{\bm{\Omega}}(s)$ denotes the Magnus exponent derived from the strain twist using the second-order Lie-group integration method described in \cite{renda2020geometric}.

The coupled Jacobian matrix of the cross-section, expressed in the local frame $O_s$, defines the mapping from the generalized joint-coordinate rates to the twist of the TD-ACM cross-section at arc-length $s$:
\begin{equation}
\bm{J}_s^s(\bm{q},s)
= \operatorname{Ad}_{\bm g_s^a}^{-1}\operatorname{Ad}_{\bm g_a^b}^{-1}
\,\bm{J}_b^b(\bm{q})
\!+ \operatorname{Ad}_{\bm{g_s^a}}^{-1}
\int_{0}^{s}
\operatorname{Ad}_{\bm{g_s^a}}
\, \bm{B}_q(\sigma)\, d\sigma.
\label{eq:Js}
\end{equation}
where $\operatorname{Ad}_{(\cdot)}$ denotes the adjoint representation and $\bm{B}_q(\sigma) \in \mathbb{R}^{6 \times n_s}$ is the finite basis matrix mapping the CR generalized coordinates to the strain shape functions.

The tip Jacobian expressed in the local tip frame $O_L$ is obtained at $s=L$ as
\begin{equation}
\bm{J}_L^L(\bm{q}) \triangleq \bm{J}_s^s(\bm{q},L),
\end{equation}
and maps $\dot{\bm q}$ to the end-effector twist as
$\bm{v}_L^{L} = \bm{J}_L^L(\bm{q}) \dot{\bm{q}}$.

Applying D’Alembert’s principle \cite{renda2024dynamics} and modeling the UAV as a rigid body \cite{szasz2022modeling}, the governing dynamic equation of the coupled TD-ACM is
\begin{equation}
\bm{M}(\bm{q}) \ddot{\bm{q}} 
+ \left( \bm{C}(\bm{q}, \dot{\bm{q}})+ \bm{D}(\bm{q}) \right) \dot{\bm{q}} 
+ \bm{K}(\bm{q})
= \bm{\tau}_a + \bm{\tau}_{ext}(\bm{q}),
\label{equation: D5}
\end{equation}
\noindent where $\bm{\tau}_a \in \mathbb{R}^{n}$ denotes the generalized actuation vector of the coupled system. The generalized inertia and Coriolis matrices are
\begin{equation}
\bm{M}(\bm{q}) = (\bm{J}_b^b)^\top \bm{M}_b \bm{J}_b^b
+ \int_{0}^{L} (\bm{J}_s^s)^\top \bm{\bar{M}}_s \bm{J}_s^s \, ds,
\end{equation}
\begin{align}
\bm{C}(\bm{q}, \dot{\bm{q}}) =
&\ (\bm{J}_b^b)^\top \left( \operatorname{ad}^*_{\bm{v}_b^{b}} \bm{M}_b \bm{J}_b^b + \bm{M}_b \dot{\bm{J}}_b^b \right) \notag \\
&+ \int_{0}^{L} (\bm{J}_s^s)^\top \left( \operatorname{ad}^*_{\bm{v}_s^{s}} \bm{\bar{M}}_s \bm{J}_s^s + \bm{\bar{M}}_s \dot{\bm{J}}_s^s \right) \, ds,
\end{align}
where $\bm{v}_s^{s} = \bm{J}_s^s(\bm{q},s)\dot{\bm{q}}$ is the cross-section twist expressed in $O_s$, $\bm{M}_b \in \mathbb{R}^{6\times6}$ is the UAV generalized inertia expressed in $O_b$, $\bm{\bar{M}}_s \in \mathbb{R}^{6\times6}$ is the inertia density matrix of the CR cross-section expressed in $O_s$, and $\operatorname{ad}^*_{(\cdot)}$ denotes the Lie algebra co-adjoint operator.

The matrix that maps the tendon tension inputs to the generalized forces is
\begin{equation}
\bm{B}_s(\bm{q}) = \int_{0}^{L} \bm{B}_{q}^\top \bm{B}_a\, ds \in \mathbb{R}^{n_s \times n_a},
\end{equation}
where $\bm{B}_a \in \mathbb{R}^{6 \times n_a}$ maps the tendon tensions to the distributed actuation wrench acting on the CR \cite{renda2017screw}. Likewise, for an underactuated quadrotor, the matrix that maps the quadrotor inputs to the generalized forces is defined as
\begin{equation}
\bm{B}_m(\bm{q}) = \begin{bmatrix}
\bm{T}^\top & \bm{0}_{3 \times 3} \\
\bm{0}_{3 \times 3} & \bm{R}_b
\end{bmatrix}\bm{\Gamma},
\end{equation}
where
\begin{equation}
\bm{\Gamma} =
\begin{bmatrix}
0 & r_m & 0 & -r_m\\
-r_m & 0 & r_m & 0\\
k_d & -k_d & k_d & -k_d\\
0 & 0 & 0 & 0\\
0 & 0 & 0 & 0\\
1 & 1 & 1 & 1
\end{bmatrix},
\end{equation}
with $r_m$ denoting the distance from each motor to the UAV CoM and $k_d$ the drag factor. Thus, the generalized actuation matrix is
\begin{equation}
\bm{B}(\bm{q}) =
\begin{bmatrix}
\bm{B}_m(\bm{q}) & \bm{0}_{6\times n_a} \\
\bm{0}_{n_s \times 4} & \bm{B}_s(\bm{q})
\end{bmatrix}
\in \mathbb{R}^{n \times (4+n_a)}.
\end{equation}
Hence, $\bm{\tau}_a=\bm{B}(\bm{q}) \bm{\tau}$, where the actuator input vector is
\begin{equation}
\bm{\tau} = 
\begin{bmatrix} 
\bm{\tau}_m^\top & \bm{\tau}_s^\top 
\end{bmatrix}^\top 
\in \mathbb{R}^{4+n_a},
\end{equation}
comprising the quadrotor actuation vector $\bm{\tau}_m \in \mathbb{R}^{4}$ and the tendon tension inputs $\bm{\tau}_s \in \mathbb{R}^{n_a}$. Moreover, the generalized external force vector is obtained as
\begin{equation}
\bm{\tau}_{ext}(\bm{q})
=
(\bm{J}_b^b(\bm{q}))^\top \bm{W}_b^b
+
\int_{0}^{L}
(\bm{J}_s^s(\bm{q},s))^\top
\bm{\bar{W}}_{ext}(s)
\, ds,
\end{equation}
where $\bm{W}_b^b \in \mathbb{R}^{6}$ denotes the point wrench applied at the UAV CoM expressed in $O_b$, and $\bm{\bar{W}}_{ext}(s) \in \mathbb{R}^{6}$ denotes the distributed external wrench density acting on the CR cross-section at $s$, expressed in the local frame $O_s$.

Assuming zero stiffness and damping for the rigid UAV, the generalized damping matrix and elastic force vector are only contributed by the CR:
\begin{equation}
\bm{D}_s = \int_{0}^{L} \bm{B}_q^\top \bm{\bar{D}} \bm{B}_q \, ds,
\end{equation}
\begin{equation}
\bm{K}_s(\bm q) = \int_{0}^{L} \bm{B}_q^\top \bm{\bar{H}} \left( \bm{\xi} - \bm{\xi}^* \right) ds,
\end{equation}
where $\bm{\bar{D}},\ \bm{\bar{H}} \in \mathbb{R}^{6\times6}$ denote the damping and elasticity matrices, respectively, and $\bm{\xi}^*$ is the reference strain configuration. The full generalized damping matrix is given by
\begin{equation}
\bm{D} =
\begin{bmatrix}
\bm{0}_{6\times 6} & \bm{0}_{6\times n_s} \\
\bm{0}_{n_s \times 6} & \bm{D}_s
\end{bmatrix},
\end{equation}
and the generalized elastic force vector is assembled as
\begin{equation}
\bm{K}(\bm q) =
\begin{bmatrix}
\bm{0}_{6\times 1} \\
\bm{K}_s(\bm q)
\end{bmatrix}.
\end{equation}

\section{Dual-Camera Servoing Algorithm}

The proposed dual-camera image-guided algorithm employs both a local tip camera, $C_L$, and a global camera mounted on the UAV, $C_G$. The decision-making module switches between an advanced IBVS scheme using the local camera and pre-planned trajectory tracking, depending on the initial fiducial marker pose. Prior to the primary control loop, a preliminary image-processing stage is implemented to compensate for attitude-induced UAV motion in the image space.

\subsection{UAV Roll and Pitch Compensation in Image Space}

The underactuated nature of the UAV prevents direct control of its roll and pitch angles for tip positioning, as these angles are regulated by the low-level controller to generate lateral motion. In an image-guided control framework, roll and pitch rotations induce apparent motion in the onboard camera images. To compensate for this effect, we propose a formulation on $\mathrm{SE}(3)$ in which these rotations are virtually removed through the introduction of a virtual camera, denoted by $C_V$.

The homogeneous transformation of the local camera $C_L$ with respect to the inertial frame $O_I$ is given by

\begin{equation}
\bm{g}_{C_L} = \bm{g}_b \bm{g}_{a}^{b} \bm{g}_L^a \bm{g}_{C_L}^L,
\end{equation}
where $\bm{g}_L^a$ denotes the transformation of the tip frame $O_L$ with respect to the attachment frame $O_a$, and $\bm{g}_{C_L}^L$ denotes the transformation of $C_L$ with respect to $O_L$. 

To eliminate the effect of roll and pitch rotations, we define a virtual UAV pose $\bm{g}_\psi$ by setting $(\phi,\theta)=(0,0)$ while preserving the translation $\bm{r}_b$. The corresponding transformation of the virtual camera $C_V$ is

\begin{equation}
\bm{g}_{C_V} = \bm{g}_\psi \bm{g}_{a}^{b} \bm{g}_L^a \bm{g}_{C_L}^L.
\end{equation}

The relative transformation between $C_L$ and $C_V$ is

\begin{equation}
\bm{g}_{C_L}^{C_V} = \bm{g}_{C_V}^{-1} \bm{g}_{C_L}.
\end{equation}
Because the two configurations differ only by a rotation at the UAV level, the translational components cancel, yielding a pure rotation:
\begin{equation}
\bm{g}_{C_L}^{C_V} =
\begin{bmatrix}
\bm{R}_{C_L}^{C_V} & \bm{0}_{3\times1} \\
\bm{0}_{1\times3} & 1
\end{bmatrix},
\end{equation}
where $\bm{R}_{C_L}^{C_V} = \bm{R}_{C_V}^\top \bm{R}_{C_L}$.

As a result, the homogeneous coordinates of a marker point transform as
\begin{equation}
\tilde{\bm{r}}_{p}^{C_V}
=
\bm{g}_{C_L}^{C_V}
\tilde{\bm{r}}_{p}^{C_L},
\end{equation}
where
$\tilde{\bm{r}}_{p}^{C_L}, \tilde{\bm{r}}_{p}^{C_V} \in \mathbb{R}^{4}$
are the homogeneous coordinate representations of the marker expressed in frames $O_{C_L}$ and $O_{C_V}$, respectively. Let the Euclidean coordinates of the marker satisfy
\begin{equation}
\bm{r}_{p}^{C_L} = z_p^{C_L} \bm{p}_p^{C_L}, 
\qquad
\bm{r}_{p}^{C_V} = z_p^{C_V} \bm{p}_p^{C_V},
\end{equation}
where $\bm{p}_p^{C_L}$ and $\bm{p}_p^{C_V}$ denote the normalized image projections in frames $O_{C_L}$ and $O_{C_V}$, respectively. The normalized projection in the virtual camera frame becomes
\begin{equation}
\bm{p}_p^{C_V}
=
\frac{z_p^{C_L}}{z_p^{C_V}}
\bm{R}_{C_L}^{C_V}
\bm{p}_p^{C_L}.
\end{equation}

Under the assumption of negligible depth variation and small relative displacement between the actual and virtual camera frames, the compensated projection for the $n_f$ feature points can be approximated as in \cite{kim2016vision}:
\begin{equation}
\bm{p}_{p_k}^{C_V}
\approx
\bm{R}_{C_L}^{C_V}
\bm{p}_{p_k}^{C_L},
\qquad k \in \{1, \ldots, n_f\}.
\end{equation}

\subsection{IBVS in Local Camera with Advanced Features}

The IBVS proposed here offers several advantages over point-feature-based IBVS. First, it guarantees global stability. Second, it decouples translational motion from rotational motion. Third, the interaction matrix has a fixed and relatively small dimension, which facilitates real-time implementation, particularly when computing its pseudo-inverse \cite{ahmadi2021robust, keshmiri2016image}. The vector of advanced image features is defined as:

\begin{equation}
    \bm p^*  \triangleq 
    \begin{bmatrix}
        p_{\varphi}^*, \; p_{\theta}^*, \; p_{\psi}^*, \; p_x^*, \; p_y^*, \; p_z^*
    \end{bmatrix}^\top.
\end{equation}

Let $\{\bm{p}_{p_k}^{C_V}\}_{k=1}^4$, where $\bm{p}_{p_k}^{C_V} = [u_k^{C_V},\; v_k^{C_V},\; 1]^\top$, denote a set of image features in the virtual camera. The centroid of these features defines $p_x^*$ and $p_y^*$ as follows:

\begin{equation}
    p_x^* = \frac{\sum_{k=1}^4 u_k^{C_V}}{4},  p_y^* = \frac{\sum_{k=1}^4 v_k^{C_V}}{4}.
\end{equation}

\noindent The feature $p_z^*$ is defined based on the perimeter of the polygon formed by the four image features:
\begin{equation}
\begin{aligned}
     p_z^* &= \sqrt{(u_2^{C_V} - u_1^{C_V})^2 + (v_2^{C_V} - v_1^{C_V})^2} \\
    &\quad + \sqrt{(u_3^{C_V} - u_2^{C_V})^2 + (v_3^{C_V} - v_2^{C_V})^2} \\
    &\quad + \sqrt{(u_4^{C_V} - u_3^{C_V})^2 + (v_4^{C_V} - v_3^{C_V})^2} \\
    &\quad + \sqrt{(u_4^{C_V} - u_1^{C_V})^2 + (v_4^{C_V} - v_1^{C_V})^2}.
\end{aligned}
\end{equation}

The rotational components of the feature vector are obtained from the deformation of the image features induced by camera rotation about its local coordinate frame and are defined as:

\begin{equation}
p_{\varphi}^* =
\frac{1}{2}\!\left[
\tan^{-1}\!\!\left(\frac{u_2^{C_V}-u_1^{C_V}}{v_1^{C_V}-v_2^{C_V}}\right)
-
\tan^{-1}\!\!\left(\frac{u_3^{C_V}-u_4^{C_V}}{v_4^{C_V}-v_3^{C_V}}\right)
\right],
\label{eq33}
\end{equation}

\begin{equation}
p_{\theta}^* =
\frac{1}{2}\!\left[
\tan^{-1}\!\!\left(\frac{v_3^{C_V}-v_2^{C_V}}{u_3^{C_V}-u_2^{C_V}}\right)
\!-\!
\tan^{-1}\!\!\left(\frac{v_4^{C_V}-v_1^{C_V}}{u_4^{C_V}-u_1^{C_V}}\right)
\right],
\label{eq34}
\end{equation}

\begin{equation}
p_{\psi}^* =
\frac{1}{2}\!\left[
\tan^{-1}\!\!\left(\frac{v_1^{C_V}-v_4^{C_V}}{u_4^{C_V}-u_1^{C_V}}\right)
\!+\!
\tan^{-1}\!\!\left(\frac{v_2^{C_V}-v_3^{C_V}}{u_3^{C_V}-u_2^{C_V}}\right)
\right].
\label{eq35}
\end{equation}

Let $\bm{p}^*_d$ be the desired image feature vector. The image feature error vector is defined as
\begin{equation}
\bm e_{p^*} \triangleq \bm{p}^* - \bm{p}^*_d.
\end{equation}

In cases where $C_V$ is able to adequately capture the target image features, the IBVS-commanded twist is utilized to compute the required velocity of $C_V$, expressed in its local frame $O_{C_V}$, according to \cite{Corke2011, Sepahvand2025}:

\begin{equation}
\bm{v}_{C_V}^{C_V} = -\lambda \bm{J}_{\text{img}}^{-1} \bm e_{p^*},
\label{equation:39}
\end{equation}

\noindent where $\bm{J}_{\text{img}} \in \mathbb{R}^{6 \times 6}$ denotes the interaction matrix (image Jacobian), and $\lambda > 0$ is a control gain that regulates the convergence rate.

Using the fixed transformation $\bm{g}_{C_V}^{L}$ between the tip frame and the virtual camera frame, the corresponding tip twist expressed in $O_L$ is obtained as $\bm{v}_{L}^{L}
=\operatorname{Ad}_{\bm{g}_{C_V}^{L}}
\bm{v}_{C_V}^{C_V}$.
Accordingly, the desired rate of change of the generalized coordinates is computed as $\bm{\dot{q}}^{d} = \left( \bm{J}_{L}^{L}(\bm{q}) \right)^{\dagger}
\bm{v}_{L}^{L}$, where $(\cdot)^\dagger$ represents the Moore--Penrose pseudoinverse.

\subsection{Smooth Joint Trajectory Tracking}

We propose a fast and practical strategy for scenarios in which image features leave the FoV of the local camera $C_L$. In such cases, a switching module activates an auxiliary recovery strategy that drives the CR to restore the features to the FoV, enabling seamless continuation of the VS task. During this recovery phase, visual feedback from the global camera $C_G$ is exploited.

Let $\bm{p}$ denote the 3D point corresponding to the target center. Its normalized projection in the local and global camera frames are denoted by $\bm{p}_{p}^{C_L}$ and $\bm{p}_{p}^{C_G}$, respectively. Once the relative pose of the target with respect to $C_G$ is obtained, the relative pose between $C_L$ and the target point can be estimated through the coupled kinematic chain as

\begin{equation}
\tilde{\bm r}_{p}^{C_L} =
\bm{g}_L^{C_L}
\left(\bm{g}_L^a\right)^{-1}
\left(\bm{g}_{a}^{b}\right)^{-1}
\bm{g}_{C_G}^b
\begin{bmatrix}
z_p^{C_G}\,\bm{p}_p^{C_G}\\
1
\end{bmatrix},
\end{equation}
\noindent where $\bm{g}_{C_G}^b$ denotes the fixed transformation from the global camera frame to the UAV body frame $O_b$. Let $\bm r_p^{C_L}\in\mathbb{R}^3$ denote the Euclidean coordinates of the target point expressed in $O_{C_L}$, obtained as the first three components of $\tilde{\bm r}_{p}^{C_L}$. Ideally, when the target center is aligned with the image center of $C_L$, the lateral coordinates satisfy $x_p^{C_L}=y_p^{C_L}=0$ while the point remains in front of the camera. Thus, the desired CR configuration is obtained from

\begin{equation}
\bm{q}_s^d = \arg\min_{\bm{q}_s}
\left\lVert
\begin{bmatrix}
\bm a_x^\top\\[2pt]
\bm a_y^\top
\end{bmatrix}
\bm r_p^{C_L}
\right\rVert_2, \quad \bm a_z^\top \bm r_p^{C_L} > 0,
\end{equation}

\noindent where $\bm a_x=[1,0,0]^\top$, $\bm a_y=[0,1,0]^\top$, and $\bm a_z=[0,0,1]^\top$, and $\lVert\cdot\rVert_2$ denotes the Euclidean norm. The configuration obtained from this minimization positions the CR such that the local camera can re-acquire the target point. A cubic joint-space polynomial is then used to generate smooth trajectories toward $\bm q_s^d$:

\begin{equation}
q_{s_i}^d(t) = \sum_{j=0}^{3} a_{ij} t^j, 
\quad i \in \{1,\ldots,n_s\},
\end{equation}
\noindent where $a_{ij}$ are constant coefficients and $t \in [t_0,t_f]$. The trajectory satisfies boundary conditions that enforce zero initial and final velocities:

\begin{equation}
q_{s_i}^d(t_0)=q_i^0, \quad
q_{s_i}^d(t_f)=q_i^f,
\quad i \in \{1,\ldots,n_s\}.
\end{equation}

\begin{equation}
\dot q_{s_i}^d(t_0)=0, \quad
\dot q_{s_i}^d(t_f)=0,
\quad i \in \{1,\ldots,n_s\}.
\end{equation}

Finally, the reference velocity supplied to the model-based controller is defined as $\dot{\bm q}^{d} =
\begin{bmatrix}
\bm 0^\top, &
\bm 0^\top, &
(\dot{\bm q}_s^{d})^\top
\end{bmatrix}^\top$, which implies that only the CR coordinates are actively commanded during the recovery phase while the UAV pose remains unchanged.

\section{Model-Based Control}

Although the high-level image-guided control system generates the joint-space velocities required for task-space twist control, a model-based low-level controller remains beneficial. The UAV’s underactuated nature, strong nonlinear subsystem interactions, and external disturbances require a dedicated controller to ensure stable joint-space control. However, controllers assuming perfect system dynamics are unreliable for the TD-ACM due to unmodeled dynamics, particularly the difficulty in explicitly determining system inertia. In addition, the soft materials exhibit variable properties sensitive to hysteresis and temperature. Therefore, a robust and adaptive controller is proposed to mitigate model uncertainties within the control law.

\subsection{Adaptive Controller}

To ensure accurate joint trajectory tracking, the desired joint velocities, $\bm{\dot{q}}^d$, are adjusted in response to tracking errors. The proposed control law incorporates a proportional correction term, $-\bm{\beta} \bm{e}$, where $\bm{\beta} \in \mathbb{R}^{n \times n}$ is a positive definite diagonal gain matrix, and $\bm{e} \triangleq \bm{q} - \bm{q}^d$ represents the joint-space tracking error. This yields the reference joint velocity $\bm{\dot{q}}_r$, defined as 
\begin{equation}
\bm{\dot{q}}_r = \bm{\dot{q}}^d - \bm{\beta} \bm{e}.
\label{equation: S2}
\end{equation}
The sliding variable $\bm{s}_q$ is defined as the difference between the actual joint velocities and the reference joint velocities as

\begin{equation}
\bm{s}_q = \bm{\dot{q}} - \bm{\dot{q}}_r = \bm{\dot{e}} + \bm{\beta} \bm{e}.
\label{equation: S3}
\end{equation}

Based on adaptive sliding mode control principles, the control law for the TD-ACM is formulated as follows. The generalized control input $\bm{\tau}_a$ is computed as

\begin{equation}
\bm{\tau}_a=
\hat{\bm{M}} \bm{\ddot{q}}_r
+ (\hat{\bm{C}} + \hat{\bm{D}}) \bm{\dot{q}}_r
+ \hat{\bm{K}}
- \bm{A_d} \bm{\dot{e}}
- \bm{A_p} \bm{e}
+ \hat{\bm{\Delta}},
\label{equation: S4}
\end{equation}

\noindent where $\hat{\bm{M}}$, $\hat{\bm{C}}$, and $\hat{\bm{D}} \in \mathbb{R}^{n \times n}$ represent the estimated inertia, Coriolis, and damping matrices, respectively, and $\hat{\bm{K}} \in \mathbb{R}^{n}$ represents the estimated generalized elastic force vector. The matrices $\bm{A_d} \in \mathbb{R}^{n \times n}$ and $\bm{A_p} \in \mathbb{R}^{n \times n}$ are positive definite gain matrices for derivative and proportional error terms. The term $\hat{\bm{\Delta}}$ accounts for the estimation of model uncertainties due to possible unmodeled components. By substituting the designated control law \eqref{equation: S4} into the system dynamics \eqref{equation: D5}, the following relation is obtained:

\begin{equation}
\begin{aligned}
&\bm{M} \bm{\ddot{q}} + (\bm{C} + \bm{D}) \bm{\dot{q}} + \bm{K} = \\
&\hat{\bm{M}} \bm{\ddot{q}}_r + (\hat{\bm{C}} + \hat{\bm{D}}) \bm{\dot{q}}_r + \hat{\bm{K}}
- \bm{A_d} \bm{\dot{e}} - \bm{A_p} \bm{e}
+ \hat{\bm{\Delta}} + \bm{\tau}_{\text{ext}}.
\end{aligned}
\label{equation: S5}
\end{equation}

Rearranging the terms in \eqref{equation: S5}, the system dynamics can be expressed as:

\begin{align}
& (\bm{M} - \hat{\bm{M}}) \bm{\ddot{q}}
+ ((\bm{C} - \hat{\bm{C}}) + (\bm{D} - \hat{\bm{D}})) \bm{\dot{q}}
+ (\bm{K} - \hat{\bm{K}}) \nonumber \\
&= -\hat{\bm{M}} \bm{\dot{s}}_q
- (\hat{\bm{C}} + \hat{\bm{D}}) \bm{s}_q
- \bm{A_d} \bm{\dot{e}}
- \bm{A_p} \bm{e}
+ \hat{\bm{\Delta}}
+ \bm{\tau}_{\text{ext}}.
\label{equation: S6}
\end{align}

The error for each term is defined as the difference between its estimated value and the actual value. This error is denoted by the notation $\tilde{(\cdot)}$. For instance, the modeling error in the inertia matrix is expressed as $\tilde{\bm{M}} \triangleq \hat{\bm{M}} - \bm{M}$. Using this notation, the dynamics becomes:

\begin{equation}
\begin{aligned}
& -\tilde{\bm{M}} \bm{\ddot{q}}
- (\tilde{\bm{C}} + \tilde{\bm{D}}) \bm{\dot{q}}
- \tilde{\bm{K}} = \\
& -\hat{\bm{M}} \bm{\dot{s}}_q
- (\hat{\bm{C}} + \hat{\bm{D}}) \bm{s}_q
- \bm{A_d} \bm{\dot{e}}
- \bm{A_p} \bm{e}
+ \hat{\bm{\Delta}}
+ \bm{\tau}_{\text{ext}}.
\end{aligned}
\label{equation: S7}
\end{equation}

\noindent Hence, the uncertainty in the model dynamics is defined as

\begin{equation}
\bm{\Delta}
=
-\tilde{\bm{M}} \bm{\ddot{q}}
- (\tilde{\bm{C}} + \tilde{\bm{D}}) \bm{\dot{q}}
- \tilde{\bm{K}}
- \bm{\tau}_{\text{ext}}.
\label{equation: S8}
\end{equation}

Accordingly, the error in estimating the uncertainty is obtained as

\begin{equation}
\tilde{\bm{\Delta}}
=
\hat{\bm{M}} \bm{\dot{s}}_q
+ (\hat{\bm{C}} + \hat{\bm{D}}) \bm{s}_q
+ \bm{A_d} \bm{\dot{e}}
+ \bm{A_p} \bm{e},
\label{equation: S9}
\end{equation}

\noindent where $\tilde{\bm{\Delta}} = \hat{\bm{\Delta}} - \bm{\Delta}$. Assuming that the actual uncertainty of the system varies more slowly than the estimation of the uncertainty, the adaptation law is defined as

\begin{equation}
\dot{\hat{\bm{\Delta}}} = -\bm{A}_a \bm{s}_q,
\label{equation: S10}
\end{equation}

\noindent where $\bm{A}_a \in \mathbb{R}^{n \times n}$ is a positive definite adaptation gain matrix.

For an underactuated UAV, the unidirectional thrust vector prevents independent control of all 6-DoF. Specifically, lateral motion in the local $x$--$y$ plane is achieved indirectly through roll and pitch regulation. Therefore, a hierarchical control architecture is adopted, in which the outer-loop translational controller computes the desired roll and pitch angles required to realize the commanded motion. Hence, we define the total force applied at the UAV CoM as $F = \left\lVert (\tau_{a4}, \tau_{a5}, \tau_{a6}) \right\rVert_2$, where $\tau_{a4}, \tau_{a5}, \tau_{a6}$ denote the translational components of $\bm{\tau}_a$. Under near-hover conditions and small-angle assumptions for roll and pitch, the desired pitch and roll angles are given by \cite{samadikhoshkho2022vision}

\begin{equation}
    \theta^d = \tan^{-1}\!\left( 
    \frac{\tau_{a4}\cos\psi^d + \tau_{a5}\sin\psi^d}{F}
    \right),
\end{equation}

\begin{equation}
    \phi^d = \tan^{-1}\!\left( 
    \frac{\tau_{a4}\sin\psi^d - \tau_{a5}\cos\psi^d}{F}
    \right).
\end{equation}


\subsection{Stability Analysis}

Note that $\hat{\bm{M}}$ is symmetric and uniformly positive definite, and
$\dot{\hat{\bm{M}}}-2\hat{\bm{C}}$ is skew-symmetric, and let $\hat{\bm{D}}$ be a constant symmetric positive semidefinite matrix.
Moreover, assume that the lumped uncertainty $\bm{\Delta}$ in \eqref{equation: S8} is constant or slowly varying, i.e., $\dot{\bm{\Delta}}=\bm{0}$ \cite{yadav2024modular}. From \eqref{equation: S3}, we have $\dot{\bm{e}}=\bm{s}_q-\bm{\beta}\bm{e}$. Substituting $\dot{\bm{e}}$ into \eqref{equation: S9} yields the closed-loop $\bm{s}_q$ dynamics
\begin{equation}
\hat{\bm{M}}\dot{\bm{s}}_q+\hat{\bm{C}}\bm{s}_q
=-(\hat{\bm{D}}+\bm{A}_d)\bm{s}_q+(\bm{A}_d\bm{\beta}-\bm{A}_p)\bm{e}+\tilde{\bm{\Delta}}.
\label{eq:s_dyn_compact}
\end{equation}

Consider the Lyapunov candidate
\begin{equation}
V=\frac12\bm{s}_q^\top\hat{\bm{M}}\bm{s}_q
+\frac12\bm{e}^\top(\bm{\beta}\bm{A}_d+\bm{A}_p)\bm{e}
+\frac12\tilde{\bm{\Delta}}^\top\bm{\Sigma}\tilde{\bm{\Delta}},
\label{eq:V_compact}
\end{equation}
where $\bm{\Sigma}\triangleq \bm{A}_a^{-1}$.
Differentiating \eqref{eq:V_compact} with respect to time and using the skew-symmetry property yields
\begin{equation}
\dot{V}
=\bm{s}_q^\top(\hat{\bm{M}}\dot{\bm{s}}_q+\hat{\bm{C}}\bm{s}_q)
+\bm{e}^\top(\bm{\beta}\bm{A}_d+\bm{A}_p)\dot{\bm{e}}
+\tilde{\bm{\Delta}}^\top\bm{\Sigma}\dot{\tilde{\bm{\Delta}}}.
\label{eq:Vdot_compact_1}
\end{equation}

Substituting \eqref{eq:s_dyn_compact} and $\dot{\bm{e}}$ into \eqref{eq:Vdot_compact_1} gives
\begin{equation}
\begin{aligned}
\dot{V}
=& -\bm{s}_q^\top(\hat{\bm{D}}+\bm{A}_d)\bm{s}_q
   +\bm{e}^\top\bm{A}_d\bm{\beta}\bm{s}_q
   -\bm{e}^\top\bm{\beta}\bm{A}_p\bm{e} \\
 & +\bm{s}_q^\top\tilde{\bm{\Delta}}
   +\tilde{\bm{\Delta}}^\top\bm{\Sigma}\dot{\tilde{\bm{\Delta}}}.
\end{aligned}
\label{eq:Vdot_compact_2}
\end{equation}

Using the adaptation law \eqref{equation: S10}, $\dot{\hat{\bm{\Delta}}}=-\bm{A}_a\bm{s}_q$, and $\dot{\bm{\Delta}}=\bm{0}$, we obtain
$\dot{\tilde{\bm{\Delta}}}=-\bm{A}_a\bm{s}_q$.
Therefore,
\[
\bm{s}_q^\top\tilde{\bm{\Delta}}
+\tilde{\bm{\Delta}}^\top\bm{\Sigma}\dot{\tilde{\bm{\Delta}}}
=
\bm{s}_q^\top\tilde{\bm{\Delta}}
-\tilde{\bm{\Delta}}^\top\bm{A}_a^{-1}\bm{A}_a\bm{s}_q
=0.
\]
Hence,
\begin{equation}
\dot{V}
=-\bm{s}_q^\top(\hat{\bm{D}}+\bm{A}_d)\bm{s}_q
+\bm{e}^\top\bm{A}_d\bm{\beta}\bm{s}_q
-\bm{e}^\top\bm{\beta}\bm{A}_p\bm{e}.
\label{eq:Vdot_compact_3}
\end{equation}

Applying Young's inequality, for any $\epsilon\in(0,1)$,
\begin{equation}
\bm{e}^\top\bm{A}_d\bm{\beta}\bm{s}_q
\le \frac{\epsilon}{2}\bm{s}_q^\top\bm{A}_d\bm{s}_q
+\frac{1}{2\epsilon}\bm{e}^\top\bm{\beta}\bm{A}_d\bm{\beta}\bm{e}.
\label{eq:young_compact}
\end{equation}

Choose gains such that
\begin{equation}
\bm{A}_p > \frac{1}{2\epsilon}\,\bm{A}_d\bm{\beta},
\label{eq:gain_compact}
\end{equation}
which implies
$\bm{\beta}\bm{A}_p-\frac{1}{2\epsilon}\bm{\beta}\bm{A}_d\bm{\beta}>0$.
Combining \eqref{eq:Vdot_compact_3}–\eqref{eq:gain_compact} and using $-\bm{s}_q^\top\hat{\bm{D}}\bm{s}_q\le 0$ yields
\begin{equation}
\dot{V}
\le -\left(1-\frac{\epsilon}{2}\right)\bm{s}_q^\top\bm{A}_d\bm{s}_q
-\bm{e}^\top\left(\bm{\beta}\bm{A}_p-\frac{1}{2\epsilon}\bm{\beta}\bm{A}_d\bm{\beta}\right)\bm{e}
\le 0.
\label{eq:Vdot_final_compact}
\end{equation}

Hence, $\bm{s}_q$, $\bm{e}$, and $\tilde{\bm{\Delta}}$ remain bounded. Moreover,
$\int_0^\infty \|\bm{s}_q(t)\|^2 dt < \infty$ and
$\int_0^\infty \|\bm{e}(t)\|^2 dt < \infty$.
Since $\hat{\bm{M}}$ is uniformly positive definite and the right-hand side of
\eqref{eq:s_dyn_compact} is bounded, it follows that $\dot{\bm{s}}_q$ is bounded,
and therefore $\bm{s}_q$ is uniformly continuous.
By Barbalat's lemma, $\bm{s}_q(t)\to \bm{0}$ as $t\to\infty$.
Finally, from the stable error dynamics $\dot{\bm{e}}+\bm{\beta}\bm{e}=\bm{s}_q$
with $\bm{\beta}>0$ and $\bm{s}_q(t)\to\bm{0}$, it follows that
$\bm{e}(t)\to\bm{0}$ and $\dot{\bm{e}}(t)\to\bm{0}$ as $t\to\infty$.

\section{Simulation Results}

In simulations, a single-section TD-CR is mounted on a quadrotor UAV. 
The physical and geometric properties of the system are summarized in Table~\ref{tab:simulation_properties_ACM}. 
To evaluate robustness against modeling inaccuracies, the estimated dynamic matrices are generated by introducing a $10\%$ multiplicative perturbation to the nominal inertia, Coriolis, damping, and stiffness matrices, thereby modeling parametric uncertainty.

The simulated environment is created in PyBullet and includes a vertical wall (\texttt{cube\_small} URDF) with a \linebreak\(20 \times 20\,\mathrm{cm}\) fiducial marker (ID~65). The intrinsic camera parameters used in simulation are listed in Table~\ref{tab:simulation_properties_cam}. 
Visual feedback is processed using OpenCV~4.10.0 with Python~3.8.10 on Ubuntu~20.04. All simulations are executed on a workstation equipped with an Intel Core i7-12700 CPU at 2.10~GHz, 16~GB RAM, and an NVIDIA GeForce RTX~3060 GPU. 

\begin{table}[b]
    \centering
    \caption{Physical and geometric properties of the TD-ACM in the strain-based simulated model.}
    \label{tab:simulation_properties_ACM}
    \setlength{\tabcolsep}{4pt}
    \renewcommand{\arraystretch}{1.05}
    \begin{tabular}{lcc}
        \hline
        \textbf{Parameter} & \textbf{Value} & \textbf{Unit} \\
        \hline
        \multicolumn{3}{l}{\textbf{Nitinol backbone}} \\
        Length ($L_1$) & $1$ & m \\
        Diameter ($d_1$) & $2\times10^{-3}$ & m \\
        Density ($\rho_1$) & $6.45\times10^{3}$ & kg/m$^3$ \\
        Young's modulus ($E$) & $5\times10^{10}$ & Pa \\
        Poisson's ratio ($\nu$) & $0.33$ & -- \\
        Material damping ($\eta$) & $10^{2}$ & Pa$\cdot$s \\
        \hline
        \multicolumn{3}{l}{\textbf{Quadrotor}} \\
        Mass ($m_2$) & $0.7331$ & kg \\
        Principal MoI ($I_{1}$) & $2.4388\times10^{-2}$ & kg$\cdot$m$^{2}$ \\
        Principal MoI ($I_{2}$) & $2.6151\times10^{-2}$ & kg$\cdot$m$^{2}$ \\
        Principal MoI ($I_{3}$) & $2.6929\times10^{-2}$ & kg$\cdot$m$^{2}$ \\
        \hline
    \end{tabular}
\end{table}

\begin{table}[b]
    \centering
    \caption{Intrinsic parameters of the local ($C_L$) and global ($C_G$) cameras in the PyBullet simulator.}
    \label{tab:simulation_properties_cam}
    \begin{tabular}{lccc}
        \hline
        \textbf{Parameter} & \textbf{$C_L$} & \textbf{$C_G$} & \textbf{Unit} \\
        \hline
        Image resolution ($W \times H$) & $1024 \times 1024$ & $500 \times 500$ & px \\
        Vertical field of view ($\mathrm{FoV}_v$) & $60$ & $71$ & deg \\
        Horizontal field of view ($\mathrm{FoV}_h$) & $60$ & $71$ & deg \\
        Focal length ($f_x = f_y$) & $886.8$ & $350.5$ & px \\
        Principal point ($c_x,c_y$) & $(512,512)$ & $(250,250)$ & px \\
        Near / far clipping planes & $0.5 / 10$ & $0.01 / 20$ & m \\
        \hline
    \end{tabular}
\end{table}

\begin{figure}
    \centering
    \includegraphics[scale = 0.48]{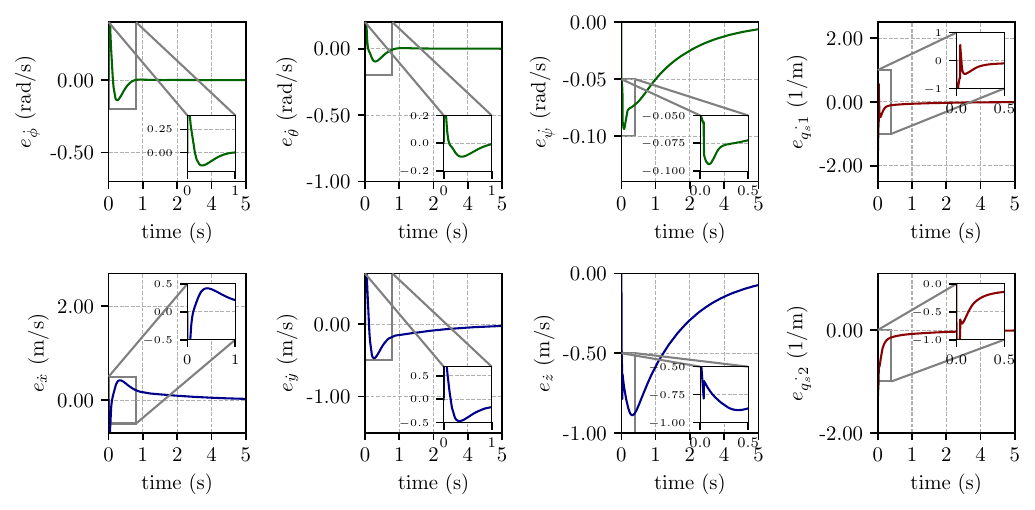}
    \caption{State error evolution with target in local camera FoV.}
    \label{fig:inside_sim_error}
\end{figure}

\begin{figure}
    \centering
    \includegraphics[scale = 0.48]{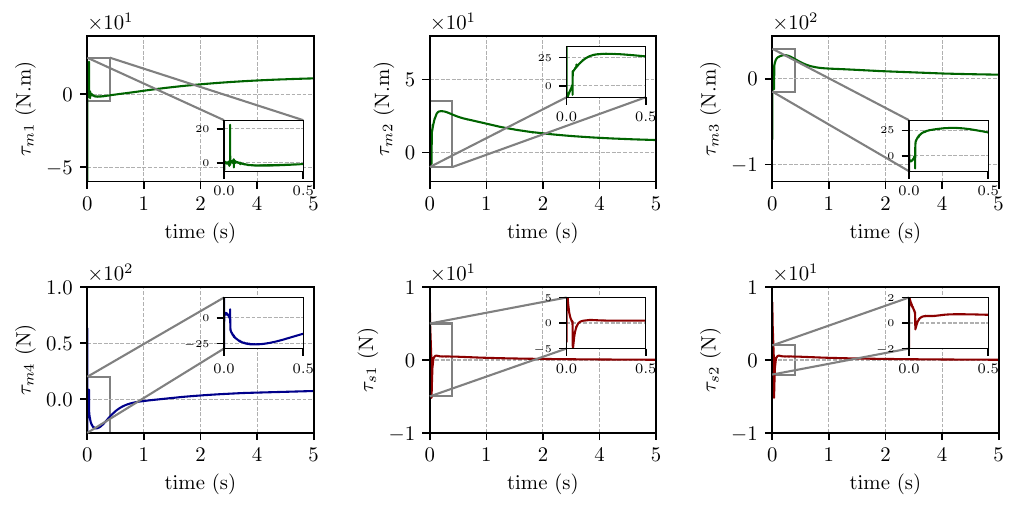}
    \caption{TD-ACM input actuation evolution with target visible to local camera.}
    \label{fig:inside_sim_torque}
\end{figure}

The initial configuration of the TD-ACM is defined by the UAV position vector
$\bm{r}_b = [-0.5,\; 0.5,\; 5]^\top\,\mathrm{m}$ and the Euler angles
$\bm{\Phi}_b = [0,\; 0,\; \pi/18]^\top\,\mathrm{rad}$, together with the strain-based configuration variables of the CR,
$\bm{q}_s = [0.3,\; 0.4]^\top$. Fig.~\ref{fig:inside_sim_error} shows that the state errors converge to zero over time as the CR tip is guided toward the desired configuration, corresponding to a desired UAV altitude of $3\,\mathrm{m}$, while all remaining generalized coordinates are regulated to zero. 
In the image space, the desired image coordinates are obtained from a reference snapshot in which the target is centered in the image captured by the local camera. 
The input actuation force applied to the underactuated coupled TD-ACM is shown in Fig.~\ref{fig:inside_sim_torque}. Although the UAV altitude and yaw angle are directly controlled, the roll and pitch angles indirectly regulate the lateral motion of the UAV. The two additional DoFs introduced by the CR compensate for the underactuation of the tip motion, resulting in a total of six independently regulated task-space DoFs.

\begin{figure}[b]
    \centering
    \includegraphics[scale = 0.4]{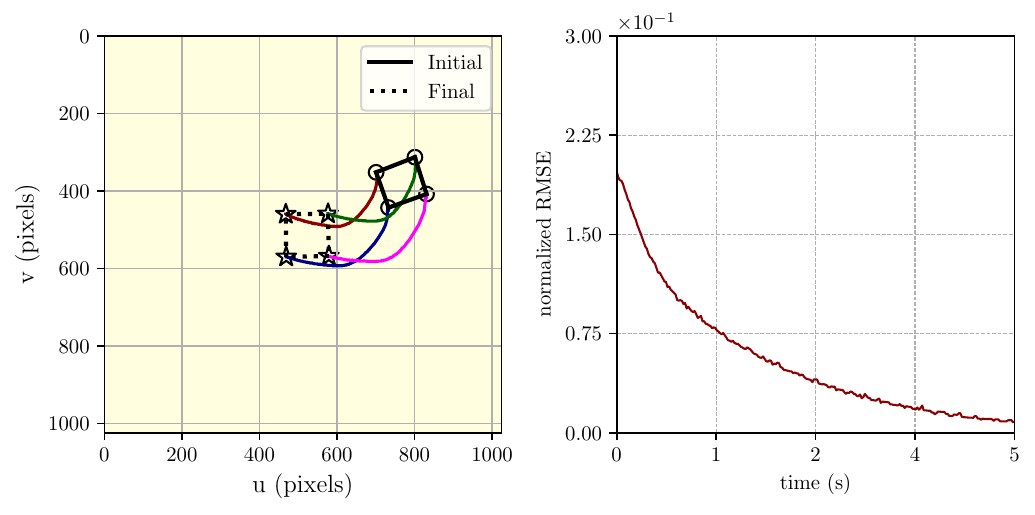}
    \caption{Feature trajectories and normalized error norm evolutions with target in local camera FoV.}
    \label{fig:inside_sim_features}
\end{figure}

For a total simulation duration of $5~\mathrm{s}$ with a time step of $0.02~\mathrm{s}$, the trajectories of the image features in the 2D image plane and the normalized root-mean-square error (RMSE) are shown in Fig.~\ref{fig:inside_sim_features}. An error growth limit of $0.05$ (normalized image-space error) is imposed for all simulations, such that if the error at any time step increases by more than $5\%$ relative to the previous step, the simulation is terminated. The initial configuration is deliberately offset in both position and orientation to demonstrate the controller’s robustness under coupled motion. For this scenario, the normalized RMSE decreases from $0.2$ to $0.01$.

\begin{figure}
    \centering
    \includegraphics[scale = 0.48]{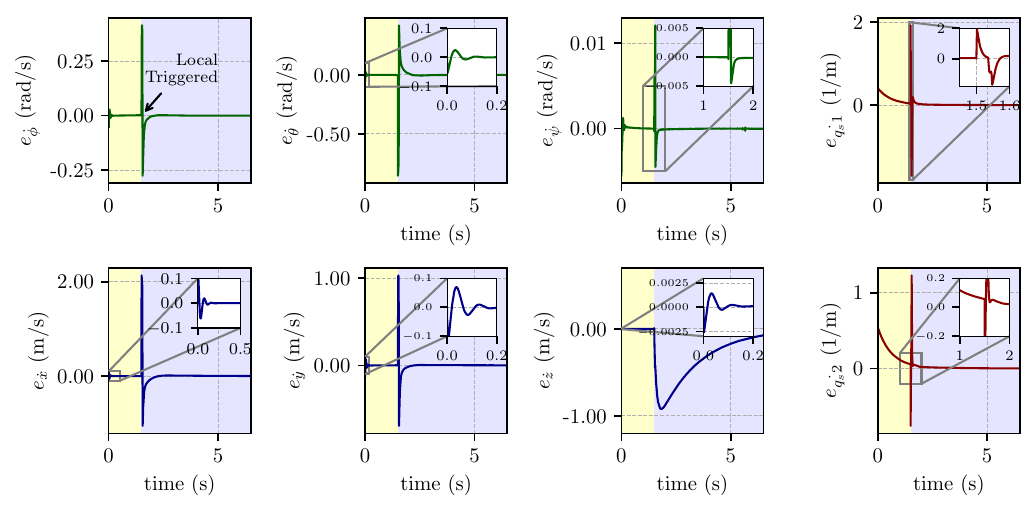}
\caption{TD-ACM state error evolution; target initially outside local camera FoV.}
    \label{fig:outside_sim_error}
\end{figure}

The second scenario evaluates the dual-camera algorithm under an out-of-FoV condition. In this configuration, the initial state of the TD-ACM is defined by
$\bm{r}_b = [0,\; 0,\; 5]^\top\,\mathrm{m}$,
$\bm{\Phi}_b = [0,\; 0,\; 0]^\top\,\mathrm{rad}$, and
$\bm{q}_s = [-0.3,\; -0.4]^\top$.
As shown in Fig.~\ref{fig:outside_sim_error}, the fiducial marker is initially outside the FoV of the local camera while remaining visible to the global camera. The motion is executed in two stages. In the first stage, the CR follows a preplanned trajectory based on a single snapshot from the global camera, while the UAV maintains a hovering condition. Once all marker features enter the FoV of the local camera, the second stage is triggered and continues until the desired configuration is achieved.

\begin{figure}
    \centering
    \includegraphics[scale = 0.48]{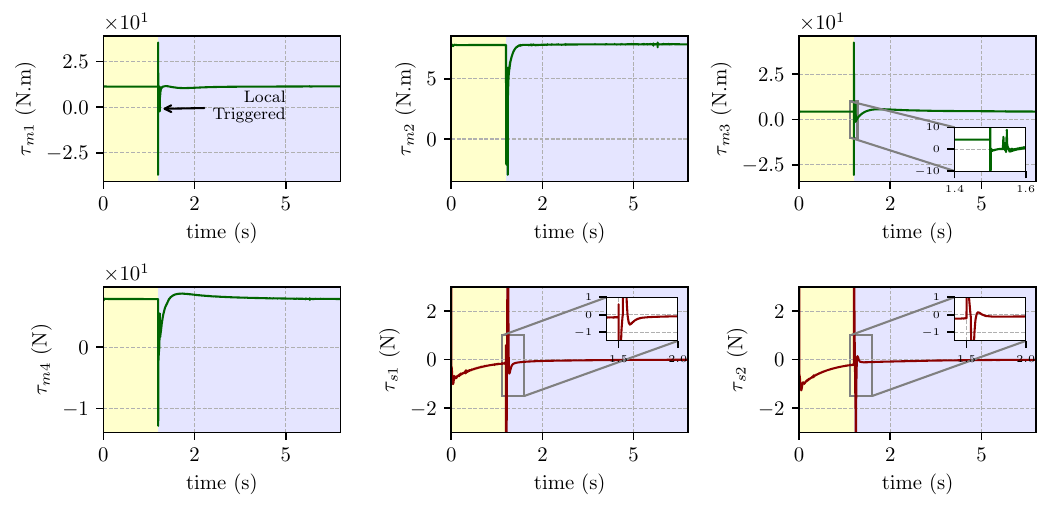}
    \caption{TD-ACM input actuation evolution; target initially outside local camera FoV.}
    \label{fig:outside_sim_tau}
\end{figure}

\begin{figure}[b]
    \centering
    \includegraphics[scale = 0.4]{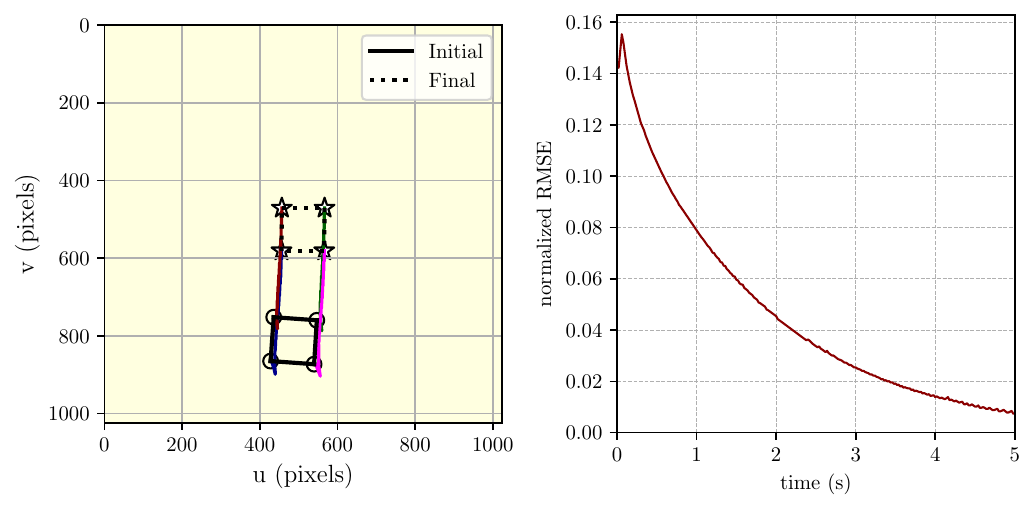}
\caption{Feature trajectories and normalized error norm evolution; target initially outside camera FoV.}
    \label{fig:outside_sim_features}
\end{figure}

Similarly, the control inputs are executed in two phases. In the first phase, the reference twists are obtained from a preplanned trajectory for the CR while the UAV maintains a hover. In the second phase, the motion is entirely driven by instantaneous images captured by the local camera to achieve precise positioning. In both phases, the actuation inputs are generated by the low-level adaptive controller in a coupled manner, as shown in Fig.~\ref{fig:outside_sim_tau}. During the second phase, the initial positions of the image features in the local camera correspond to their final positions from the first phase. The feature trajectories and the normalized RMSE are exhibited in Fig.~\ref{fig:outside_sim_features}. Despite an initial overshoot caused by the transition between phases, the features move smoothly toward their desired locations, and the RMSE is minimized. 

To provide a quantitative comparison with the coupled Lagrangian dynamic model of the TD-ACMs based on the CC assumption presented in~\cite{samadikhoshkho2021coupled}, both approaches were implemented on the same workstation under equivalent numerical settings (i.e., the same solver and time step). The computational time was measured in the simulation environment. Under the same physical parameters and initial conditions, the average computational time per iteration of the proposed dynamic formulation was $0.0339 \pm 0.003$~s, whereas the method in~\cite{samadikhoshkho2021coupled} required $0.0460 \pm 0.004$~s per iteration. These results demonstrate improved computational efficiency while simultaneously enhancing modeling accuracy. Furthermore, the proposed approach avoids computationally intensive symbolic derivations required by the existing method.

\section{Experimental Validations}

\begin{figure}[t!]
    \centering
    \includegraphics[scale = 0.40]{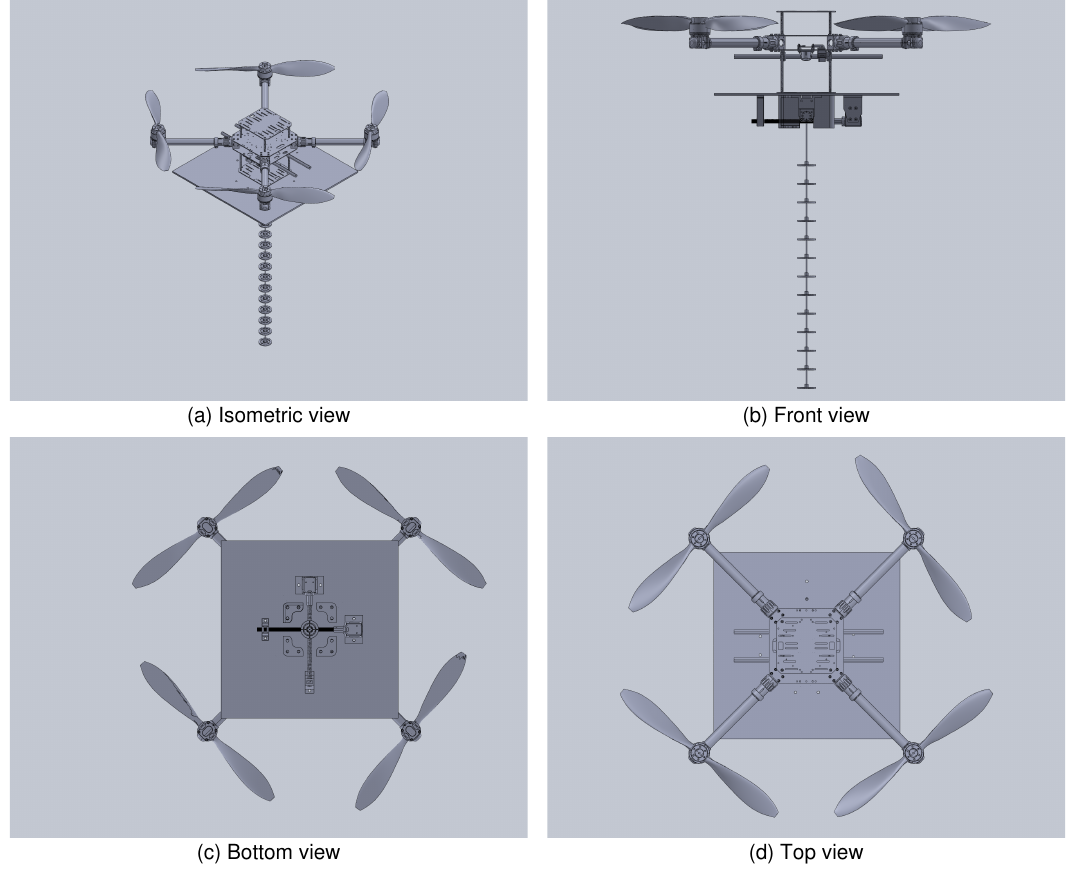}
    \caption{TD-ACM CAD model designed in SolidWorks.}
    \label{fig:setup_CAD}
\end{figure}

\begin{figure}[t!]
    \centering
    \includegraphics[scale = 0.27]{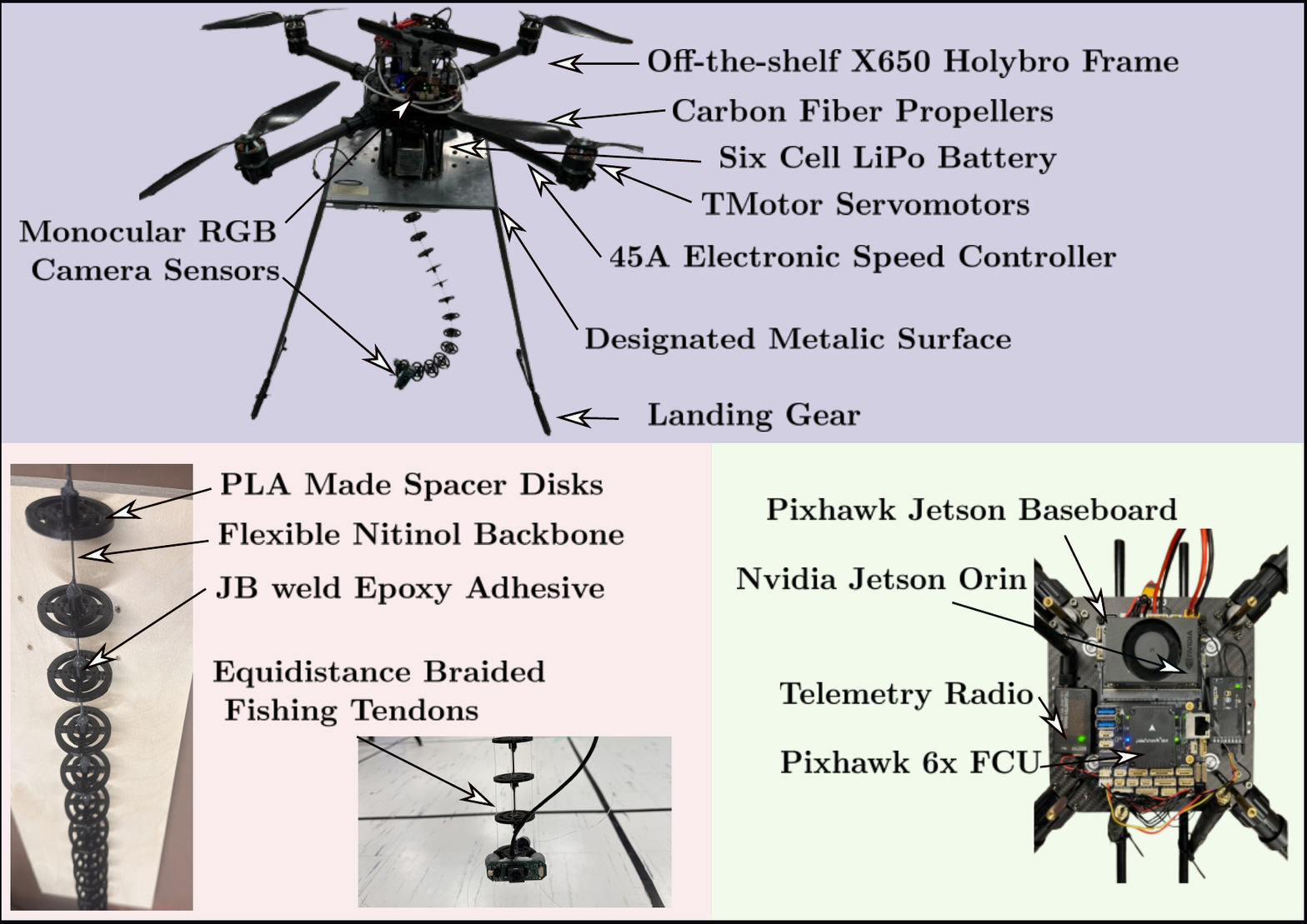}
    \caption{The customized TD-ACM experimental setup and its components.}
    \label{fig:setup_detail}
\end{figure}

A customized prototype is developed to experimentally evaluate the system performance in both ground-based and aerial scenarios. 
Fig.~\ref{fig:setup_CAD} illustrates the complete assembly of the prototype designed in SolidWorks, including the custom mounting structure and the CR. 
The main components of the experimental setup are illustrated in detail in Fig.~\ref{fig:setup_detail}.

The single-section CR consists of a Nitinol backbone actuated by four parallel tendons arranged in two symmetric pairs, enabling bending in two independent directions. 
Parallel tendon routing is achieved utilizing equidistant spacer disks along the backbone. 
The CR is mounted beneath an off-the-shelf Holybro X650 UAV frame with a fixed offset relative to the UAV CoM. 
The local camera is attached to the final disk at the CR tip, while the global camera is rigidly mounted on the UAV frame.

The system is evaluated through three series of experiments. In all tests, four AprilTags with a side length of $3~\mathrm{cm}$ are positioned at the corners of an $18 \times 12~\mathrm{cm}$ rectangular pattern mounted on a static planar object. The markers are detected using the AprilTag package \cite{Wang2016} and transmitted through a robot operating system (ROS)~Noetic middleware. Camera calibration and tag-center extraction are performed using the Kalibr package \cite{Rehder2016}. The ground control station (GCS) is an Alienware Aurora Ryzen equipped with an AMD Ryzen~9~5900 12-core processor and an NVIDIA GeForce RTX~3060 GPU. In Tests~A and~B, the UAV remains stationary and only the fixed-base CR is evaluated. In Test~C, the complete TD-ACM is evaluated as a fully coupled aerial system.

\begin{figure}[t]
    \centering

    \subfloat[]{%
        \includegraphics[width=0.47\linewidth]{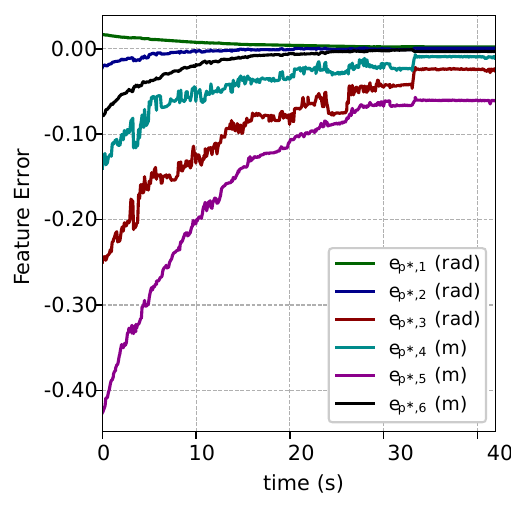}
        \label{fig:TestA_a}
    }\hfill
    \subfloat[]{%
        \includegraphics[width=0.47\linewidth]{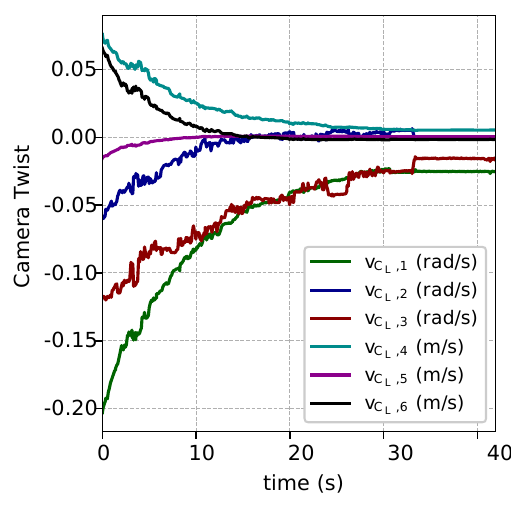}
        \label{fig:TestA_b}
    }

    \subfloat[]{%
        \includegraphics[width=0.47\linewidth]{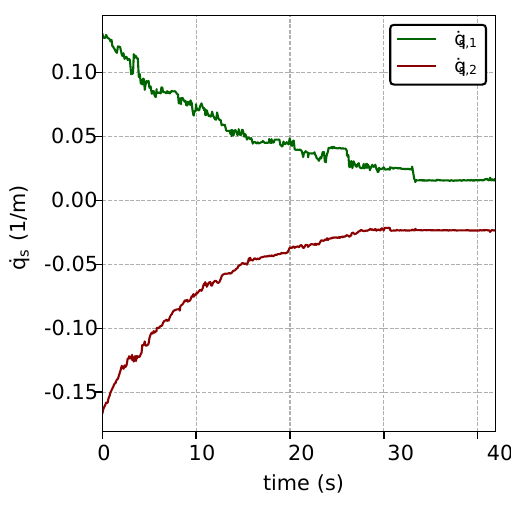}
        \label{fig:TestA_c}
    }\hfill
    \subfloat[]{%
        \raisebox{0.3cm}{\includegraphics[width=0.47\linewidth]{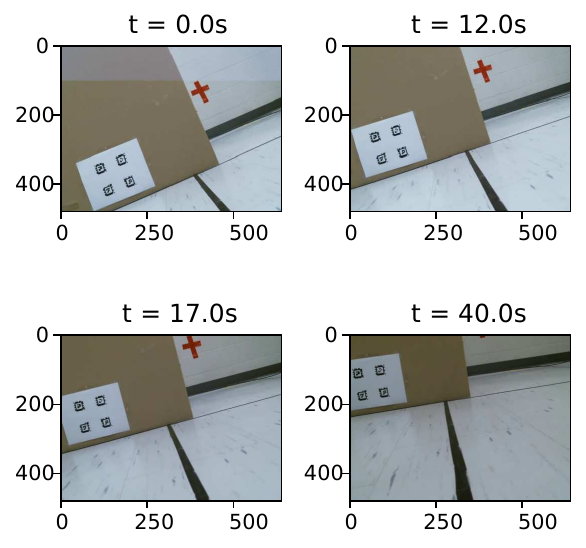}}
        \label{fig:TestA_d}
    }
    
\caption{Test A: (a) Time evolution of the advanced feature error; (b) evolution of the local camera twist; (c) time derivative of the CR generalized coordinates; and (d) four snapshots of the markers during operation.}

\label{fig:TestA}
\end{figure}

In Test~A, the marker is initially located within the FoV of the local camera, as shown in Fig.~\ref{fig:TestA_d}. The CR tip is initially misaligned relative to the desired pose, which is defined by a reference image where the feature markers appear in the top-left region of the image plane. Although the system is underactuated in this test, the initial and desired CR end-effector configurations are sufficiently close and lie within the feasible workspace. Consequently, the CR is able to drive the tip toward the desired pose. The advanced feature error decreases monotonically over time, as shown in Fig.~\ref{fig:TestA_a}, indicating convergence toward the desired state. Correspondingly, the commanded camera twists gradually converge to zero (Fig.~\ref{fig:TestA_b}), demonstrating stable velocity regulation. The strain-based generalized coordinates of the CR evolve smoothly with a decreasing rate of change, as illustrated in Fig.~\ref{fig:TestA_c}, confirming physically consistent motion.

\begin{figure}[t]
    \centering

    \subfloat[]{%
        \includegraphics[width=0.47\linewidth]{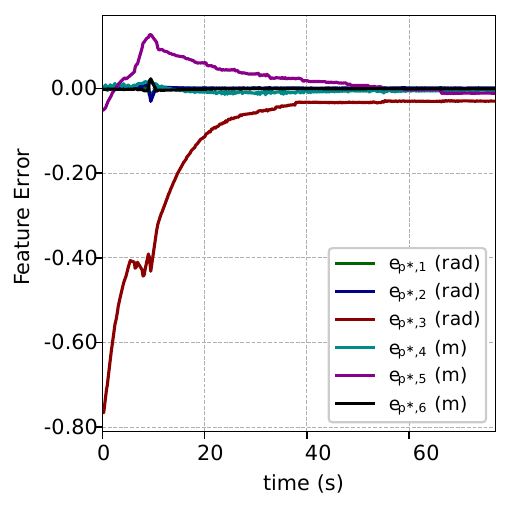}
        \label{fig:TestB_a}
    }\hfill
    \subfloat[]{%
        \includegraphics[width=0.47\linewidth]{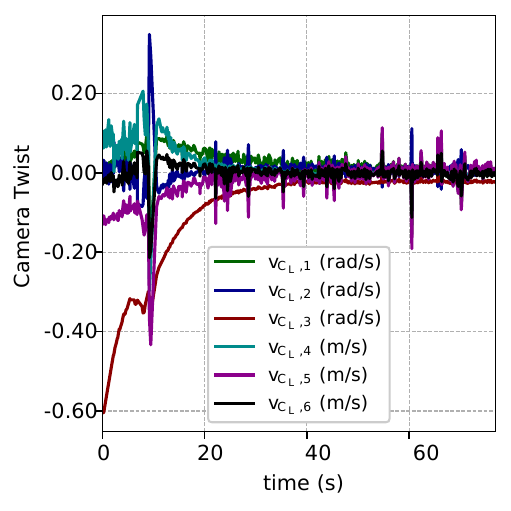}
        \label{fig:TestB_b}
    }

    \subfloat[]{%
        \includegraphics[width=0.47\linewidth]{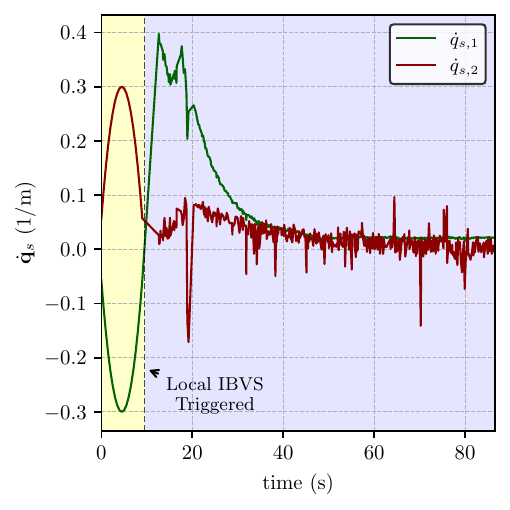}
        \label{fig:TestB_c}
    }\hfill
    \subfloat[]{%
        \raisebox{0.3cm}{%
            \includegraphics[width=0.47\linewidth, trim={0cm 0cm 0cm 0cm}, clip]{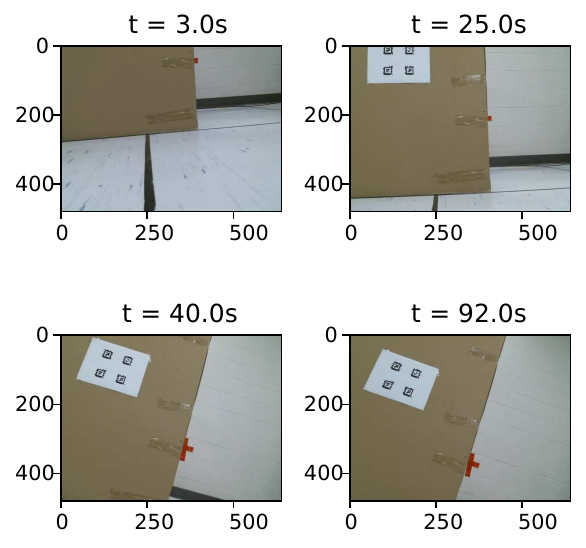}
        }
        \label{fig:TestB_d}
    }

\caption{Test B: (a) Time evolution of the advanced feature error; (b) time evolution of the local camera twist; (c) time derivative of the CR generalized coordinates; and (d) four snapshots of the markers during operation.}

\end{figure}

In Test~B, the dual-camera algorithm is evaluated on the CR. The system is initialized such that the marker lies outside the FoV of the local camera, while remaining visible to the global camera. As shown in Fig.~\ref{fig:TestB_d}, the CR is first actuated to restore the image features into the FoV of the local camera. Once detected, the features are subsequently driven toward their desired locations in the image plane. In this test, a single snapshot acquired by the global camera is used to generate the reference trajectory for the recovery phase. The planned trajectory does not need to be fully completed before switching control modes. As soon as all image features enter the FoV of the local camera, the advanced IBVS controller is activated to enable precise positioning. During the second stage, the advanced features are actively tracked and the corresponding feature errors decrease over time, as shown in Fig.~\ref{fig:TestB_a}. The associated camera twist converges toward zero (Fig.~\ref{fig:TestB_b}), indicating stable regulation. The transition between the two motion stages is clearly illustrated in Fig.~\ref{fig:TestB_c}. After approximately $16~\mathrm{s}$, the advanced IBVS is engaged, and the CR motion is entirely driven by instantaneous image feedback. In contrast to the trajectory-tracking phase, the commanded twists asymptotically approach zero toward the end of the experiment.

Test~C represents the most comprehensive experiment. Unlike the previous tests, the TD-ACM operates as a fully actuated coupled system, with interactions among all subsystems explicitly considered. The system is initialized with a large deviation, such that actuating the CR alone is insufficient to reach the desired configuration without inducing significant UAV motion. Consequently, all subsystems operate simultaneously, and their coupled interactions are clearly observable.

\begin{figure}[!ht]
    \centering

    \subfloat[]{%
        \includegraphics[width=0.47\linewidth]{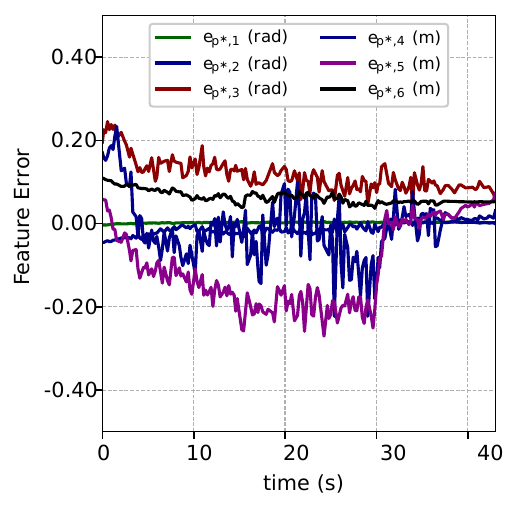}
        \label{fig:TestC_a}
    }\hfill
    \subfloat[]{%
        \includegraphics[width=0.47\linewidth]{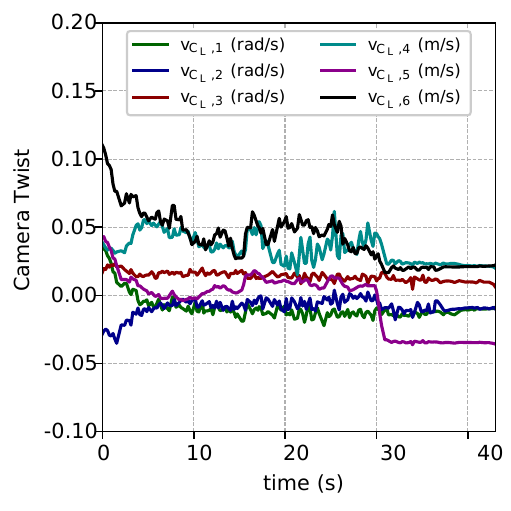}
        \label{fig:TestC_b}
    }

    \subfloat[]{%
        \includegraphics[width=0.47\linewidth]{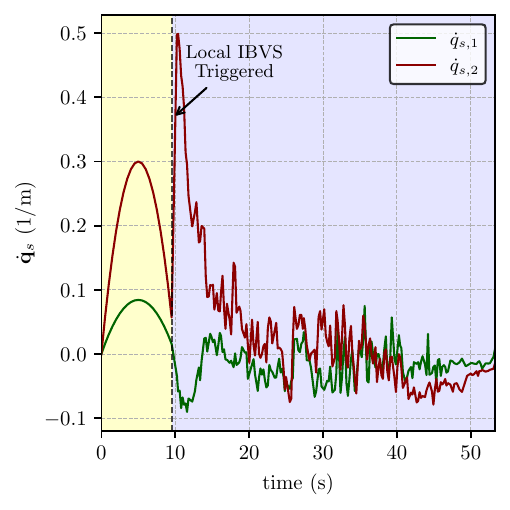}
        \label{fig:TestC_c}
    }

\caption{Test C: (a) Time evolution of the advanced feature error; (b) time evolution of the local camera twist; and (c) time derivative of the CR generalized coordinates.}

\end{figure}

We first analyze the time evolution of the advanced features. As shown in Fig.~\ref{fig:TestC_a}, oscillations in the image features are more pronounced than in the ground-based tests. This behavior arises from environmental disturbances, including structural vibrations and UAV downwash, as well as strong dynamic coupling between the UAV and the CR. The coupling effects are particularly significant during rapid UAV maneuvers, which require compensatory CR motion in directions not aligned with the final desired configuration. Despite these challenges, the feature errors are progressively reduced and eventually stabilize. The corresponding local camera twists (Fig.~\ref{fig:TestC_b}) decrease over time, indicating that the TD-ACM decelerates as it approaches the desired state. Yet, since the commanded local camera twists are coupled with the UAV motion, their final values stabilize rather than converge to zero. This is expected, as the TD-ACM maintains steady flight while preserving the desired configuration.

Fig.~\ref{fig:TestC_c} illustrates the CR twists throughout the experiment. Based on the reference image captured by the local camera, the required motion associated with the first generalized coordinate is smaller than that of the second, as the planned trajectory involves a smaller extremum along this direction. The IBVS stage exhibits higher noise compared to the ground-based tests, which is consistent with the aerial disturbances and dynamic coupling effects. Nevertheless, as the system approaches the desired pose, the CR coordinates stabilize and converge to values close to zero. This indicates that, unlike the UAV, which actively maintains a hovering condition, the CR remains essentially stationary in steady state.


\section{Conclusion}

In this work, the coupled dynamics of a TD-ACM are formulated on $\mathrm{SE}(3)$. Using D’Alembert’s principle, a Lagrangian dynamic model is derived that incorporates an actuation map projecting the motor wrenches of the underactuated UAV into the generalized wrench space. Under identical simulation and hardware conditions, the proposed formulation achieves an average $26.3\%$ reduction in computational time compared to the baseline. Based on this model, a hierarchical control architecture is developed, consisting of a high-level IBVS controller that maintains visual features within the FoV and compensates for attitude-induced image motion caused by UAV underactuation, and a low-level adaptive controller that exploits the coupled dynamics to track joint commands while accommodating lumped uncertainties. Robustness is evaluated through simulation studies with a $10\%$ multiplicative perturbation in the dynamic matrices, demonstrating convergence of the image features and improved tracking performance. The effectiveness of the proposed framework is further validated experimentally using a custom-built prototype. The aerial experiments exhibit more pronounced oscillations than the ground-based tests due to structural vibrations, UAV downwash, and strong dynamic coupling; nevertheless, the image feature error exhibits a decreasing trend. Future work will investigate aerodynamic effects, physical interaction with the environment, and extension to dual TD-CR configurations to enhance manipulation capability.

\bibliography{ref}
\bibliographystyle{IEEEtran}

\vfill
\end{document}